\documentclass[runningheads]{llncs}

\usepackage{booktabs}
\usepackage{tabularx}
\usepackage{makecell}
\usepackage{multirow}
\usepackage{siunitx}
\usepackage[super]{nth}
\usepackage[pagebackref,breaklinks,colorlinks,citecolor=blue]{hyperref}
\usepackage{cleveref}
\usepackage{orcidlink}
\usepackage{accvabbrv}

\usepackage[T1]{fontenc}
\usepackage{graphicx}
\usepackage{color}

\begin{document}

\title{Recognizing Artistic Style of Archaeological Image Fragments Using Deep Style Extrapolation}

\titlerunning{Recognizing Artistic Style of Archaeological Image Fragments}

\author{Gur Elkin\orcidlink{0009-0001-0464-7712} \and
Ofir Itzhak Shahar\orcidlink{0009-0001-3688-1159} \and
Yaniv Ohayon\orcidlink{0009-0009-1141-1909} \and
Nadav Alali\orcidlink{0009-0006-0627-5963} \and
Ohad Ben-Shahar\orcidlink{0000-0001-5346-152X}}

\authorrunning{G. Elkin et al.}

\institute{Ben-Gurion University of the Negev \\
\email{\{gurshal, shofir, yanivoha, nadavala\}@post.bgu.ac.il, obs@bgu.ac.il}}

\maketitle

\begin{abstract}
Ancient artworks obtained in archaeological excavations usually suffer from a certain degree of fragmentation and physical degradation. Often, fragments of multiple artifacts from different periods or artistic styles could be found on the same site. With each fragment containing only partial information about its source, and pieces from different objects being mixed, categorizing broken artifacts based on their visual cues could be a challenging task, even for professionals. As classification is a common function of many machine learning models, the power of modern architectures can be harnessed for efficient and accurate fragment classification. In this work, we present a generalized deep-learning framework for predicting the artistic style of image fragments, achieving state-of-the-art results for pieces with varying styles and geometries.

\end{abstract}

\section{Introduction}
\label{sec:intro}

The ability to automatically recognize and classify artistic styles from images is a challenging problem in computer vision and digital art analysis. Artistic styles encompass the distinctive visual patterns, techniques, and movements that characterize the works of different artists, periods, and schools throughout history~\cite{willats2005defining}. Accurately identifying these styles holds significant value for applications such as archiving and cataloging art collections, supporting art education and appreciation~\cite{wallraven2009categorizing}, and enabling content-based image retrieval and recommendation systems~\cite{chawla2021leveraging}.

Categorizing artistic style becomes even more complex when dealing with fragments of artistic artifacts, where only partial information is available. In many real-world scenarios, such as analyzing damaged artworks or examining details within larger compositions, the system must be capable of recognizing styles from incomplete visual data. This requirement poses unique challenges, as the absence of contextual cues and the potential loss of distinctive stylistic elements can hinder accurate classification.
In addition to partial information, physical degradation processes such as erosion, wear, fading, and discolorization, all impair the ability to correctly identify the visual style of ancient artworks.  Since clear documentation is often lacking, adequate conservation and restoration efforts become extremely challenging and necessitate advanced methods. 

One particularly important example of fragment style classification relates to frescos or cultural heritage wall paintings from different eras. Natural events such as volcanic eruptions, earthquakes or floods, as well as human destructive actions over the centuries, have damaged many frescos worldwide, reducing them to a disordered collection of small, irregularly shaped, and eroded fragments with faded textures~\cite{repair2024}. Often some pieces are missing altogether while in other cases fragments from different frescos (\ie, walls)  were mixed by natural or man-made events. During reconstruction efforts, it is thus crucial to re-sort the fragments according to their source. Hence, classifying their artistic style is critical for reconstruction, and in this paper, we indeed focus on this challenge.

Few studies have addressed the challenge of recognizing the style of fresco fragments computationally~\cite{cascone2022automatic,cascone2023classification}. These applications achieve fair accuracy using traditional machine-learning approaches based on manually crafted pictorial features that are extracted from the fragments' images. In this work, we push the state-of-the-art (SOTA) in this domain by leveraging modern deep-learning architectures next to suitable domain-specific image processing.
We present two main contributions:
\begin{enumerate}
    \item We propose a novel deep-learning architecture\footnote{See implementation at \href{https://github.com/ICVL-BGU/Fragment-Style-Recognition}{https://github.com/ICVL-BGU/Fragment-Style-Recognition}} for recognizing the artistic style of fragments, achieving SOTA results over several benchmarks.
    \item We introduce a new dataset\footnote{Download the POMPAAF dataset at \href{https://tinyurl.com/ynwc6ymm}{https://tinyurl.com/ynwc6ymm}} for fragment style classification derived from a real-world scenario. The images in this dataset were synthesized using several different fragmentation algorithms to test the effect of fragment geometry on classification accuracy.
\end{enumerate}

\section{Related Work}
\label{sec:related}

Recent advancements in deep learning methodologies have revolutionized many tasks within the field of computer vision. Among these, general style recognition and style transfer are directly related to our original contribution, where the latter is used to serve the former. Both topics are reviewed below.

\subsection{Style Recognition of Fragments}
Image classification is a central task in the field of computer vision, with deep learning approaches achieving remarkable success in recent years. Convolutional Neural Networks (CNNs) have become the de facto standard for image classification tasks, with architectures like ResNet \cite{he2016deep}, VGGNet \cite{simonyan2014very}, EfficientNet \cite{tan2019efficientnet}, and Inception \cite{szegedy2015going} demonstrating impressive performance on large-scale datasets.

In the domain of style classification, which focuses on recognizing artistic or visual styles rather than object categories, several approaches have been proposed. Karayev \etal \cite{karayev2013recognizing} pioneered the use of deep features for style recognition, showing that CNN-based features outperform traditional hand-crafted ones. Subsequent work has explored fine-tuning of pre-trained CNNs for style classification tasks, with studies demonstrating the effectiveness of transfer learning from object recognition to style recognition \cite{lecoutre2017recognizing}.

Classifying the fragments of broken artworks based on features like color descriptors or textural patterns has been studied extensively too \cite{rasheed2020classification,yang2021classification,smith2010classification}. Yet, recognizing their style as a whole was first introduced as a computational problem by Cascone \etal \cite{cascone2022automatic} who sought to distinguish fragments from the DAFNE dataset~\cite{dondi2020dafne} to 2 or 3 different styles. There, a style category was based on the artist's identity or century of creation. CNN-based approaches as well as more traditional models like random forest \cite{breiman2001random} proved competent for completing this task with high accuracy. More recently, Cascone \etal \cite{cascone2023classification} proposed the contemporary CLEOPATRA dataset with fragments of arbitrary shapes from artworks of 11 different historical eras. Naturally, the more numerous classes and the greater variance within each class resulted in lower accuracy scores by the same classification methods.

\subsection{Style Transfer}

Style transfer is a technique in computer vision and graphics that aims to render the content of one image in the style of another. Gatys \etal \cite{gatys2016image} demonstrated that deep neural networks could be leveraged to separate and recombine the content and style of arbitrary images, enabling artistic style transfer. This ability relies on the promise that the desired output could be found by optimizing a trade-off between two loss functions---the content loss and the style loss.

Content loss measures the discrepancy between the content representation of the generated image $I$ and the content image $C$, typically captured by a deeper activation $\ell$ of a CNN:
\begin{equation}
    \mathcal{L}_\mathrm{C} = \frac{1}{2}\Vert I_\ell-C_\ell\Vert_F^2
    \label{eq:content_loss_orig}
\end{equation}
Where $\Vert\cdot\Vert_F$ denotes the Forbenius norm.

On the other hand, style loss quantifies the difference between the style representation of the generated image $I$ and the style image $S$. This is achieved by comparing Gram matrices of activations from multiple layers of the CNN:
\begin{equation}
    \mathcal{L}_\mathrm{S}=\sum_\ell \frac{w_\ell}{4m_\ell^2n_\ell^2}\Vert\mathrm{Gram}(I_\ell)-\mathrm{Gram}(S_\ell)\Vert_F^2
    \label{eq:style_loss_orig}
\end{equation}
Here $m_\ell$ marks the number of feature maps at the $\ell$-th layer, and $n_\ell$ represents the product of their height and width. The contribution of each layer to the overall style loss is weighted by $w_\ell$. The Gram matrix: $\mathrm{Gram}(A) = A^\top A$ captures the correlation between each component of the latent image representations while disregarding spatial composition. This enables the style loss to embody similarity in patterns, colors, and other stylistic features between two images.

\section{Data}
\label{sec:data}

As would be further elaborated in \cref{sec:method}, our proposed approach is invariant both to the amount of different artistic styles an image fragment could be classified as, as well as to the fragment's general shape. To demonstrate that statement, besides testing our approach on CLEOPATRA dataset \cite{cascone2023classification}, we present the \textbf{Pomp}eii \textbf{A}rchive \textbf{A}rtistic-styles \textbf{F}ragments (POMPAAF) Dataset, a novel dataset containing high-quality images of Pomepian frescos, artificially broken into fragments of different sizes and shapes, all annotated with their expert-verified Pompeian artistic style.

In the following, we elaborate both on the uniqueness of the ancient Pompeian art styles as an interesting and particularly challenging test case for artistic style classification, and on the different methods used to \textit{break} the full fresco images into fragments. All fresco images originate from the Pompeii archives \cite{napolitano2023a}.

\subsection{The Four Artistic Styles of Ancient Roman Wall Paintings}
\label{sec:four_styles}
The ancient city of Pompeii, a once thriving Roman community, was buried under several meters of volcanic ash after the eruption of Mount Vesuvius in 79 AD \cite{zanker1998}. Given the unique cause of destruction, local art pieces, including frescos, mosaics, and other artifacts, were preserved in remarkably good shape, allowing modern researchers to gain valuable insights into the culture and civilization of that time, including religious beliefs, domestic environments, and public life \cite{clarke1991,zanker1998,lorenz2005,mau1882,mau1902,ling1991,berry2013,ambler2018}. 

\begin{figure}
    \centering
    \begin{tabular}{cccc}
        \includegraphics[width=0.22\columnwidth]{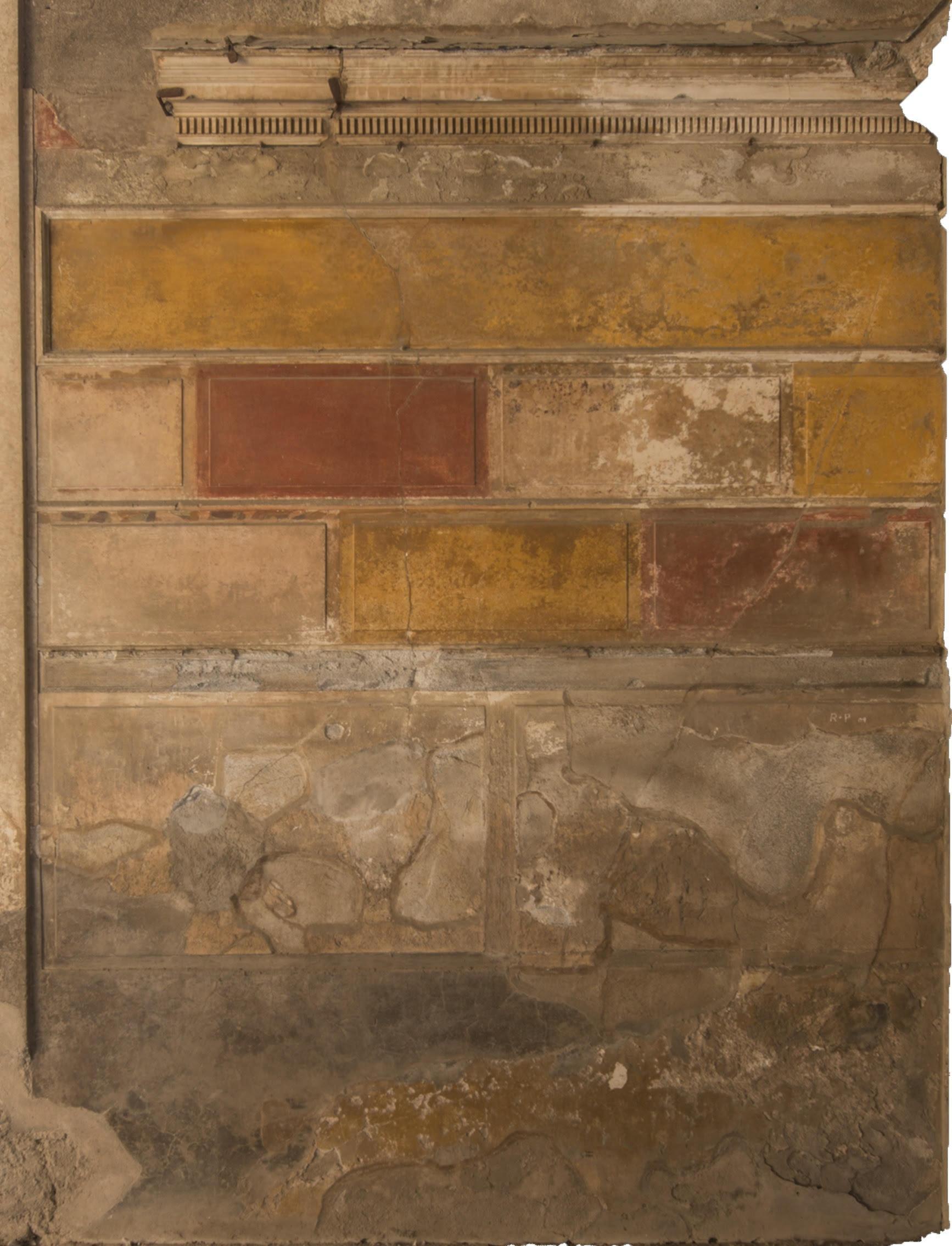}
        &
        \includegraphics[width=0.22\columnwidth]{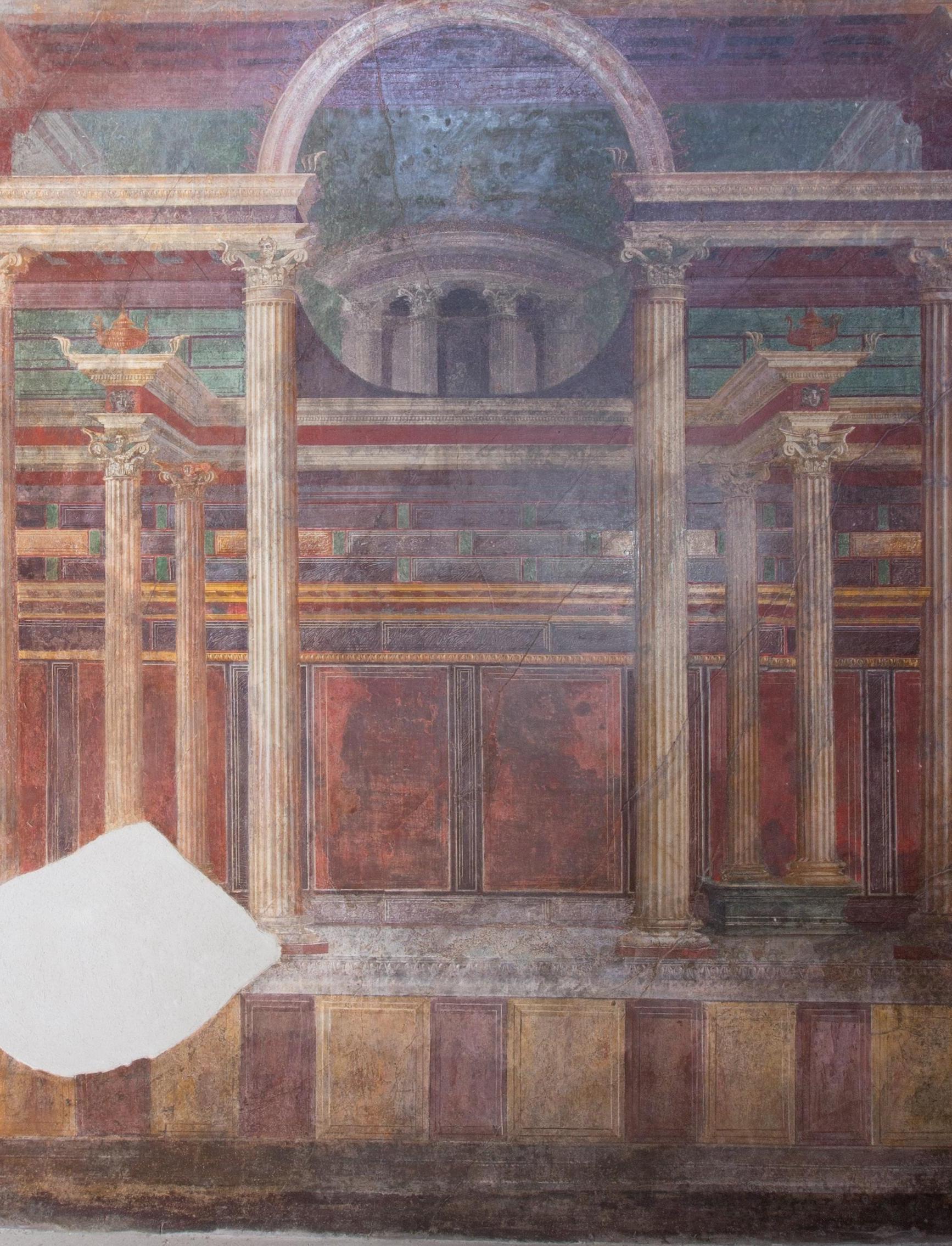}
        &
        \includegraphics[width=0.22\columnwidth]{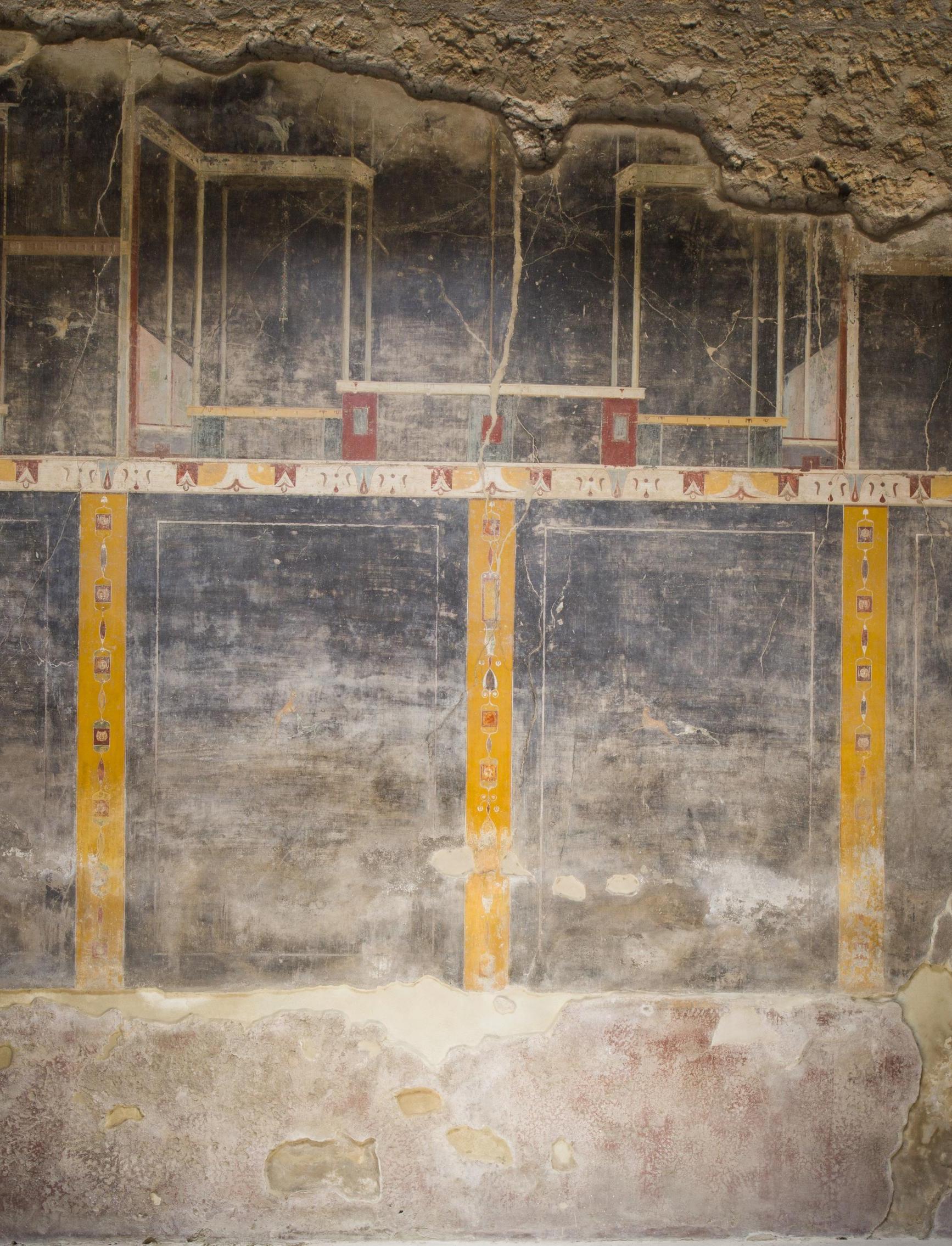}
        &
        \includegraphics[width=0.22\columnwidth]{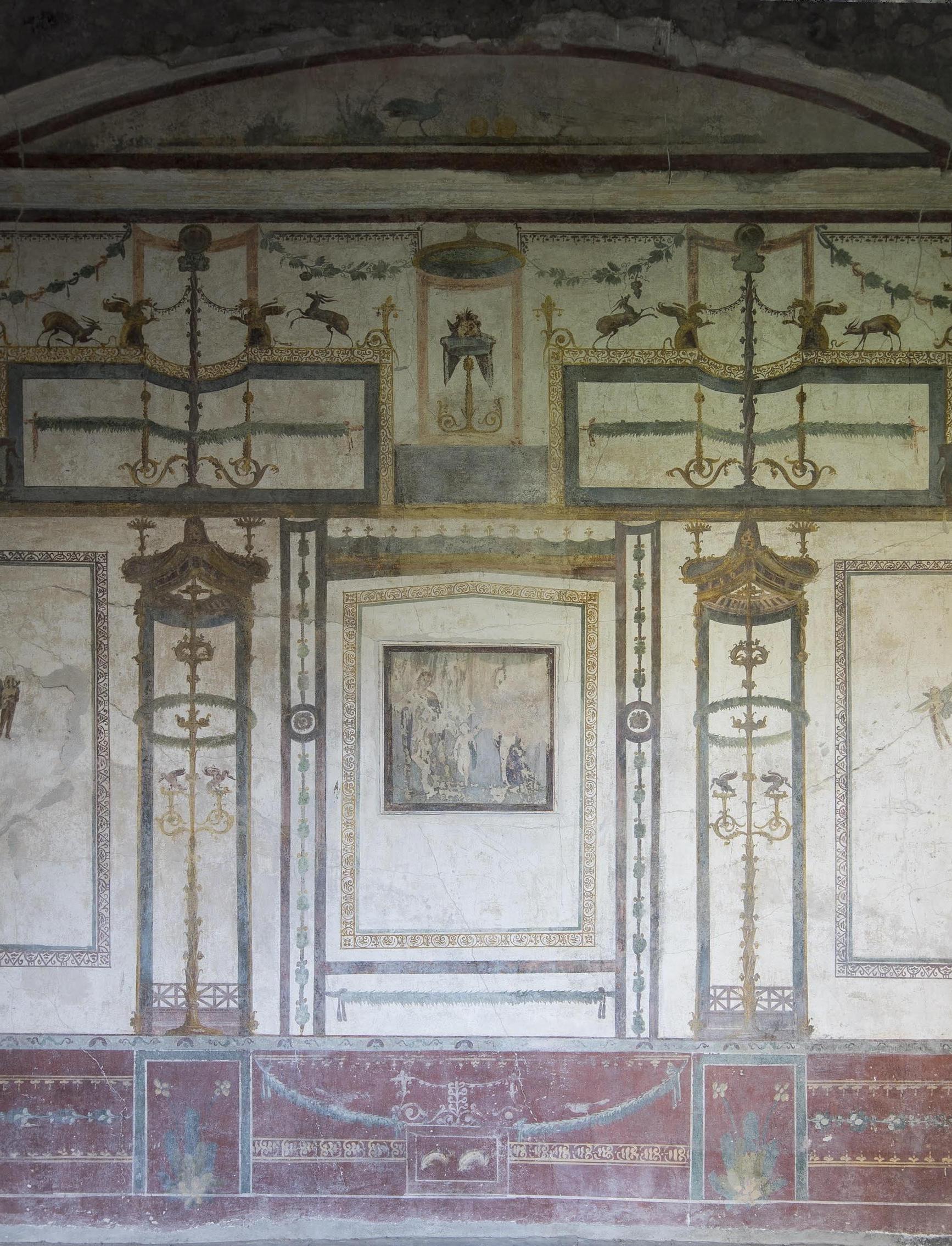} \\
        \nth{1} Style & \nth{2} Style & \nth{3} Style & \nth{4} Style
        \end{tabular}
    \caption{
           Frescos from Pompeii, representing the four styles of ancient Roman wall paintings, sampled from the Pompeii Archives published by Napolitano \etal ~\cite{napolitano2023a}.
           }
    \label{fig::styles}
\end{figure}

In 1882, the German archaeologist and art historian August Mau first categorized the wall paintings of Pompeii into four distinct styles, each associated with its own time period and artistic characteristics \cite{mau1882}:
\begin{itemize}
    \item \textbf{\nth{1} Style}, or \textit{Structural/Incrustation} style, is dated to around 200-80 BC and is believed to have originated in the Hellenistic period. This style features colorful, often three-dimensional blocks, which aim to imitate the appearance of marble.
    \item \textbf{\nth{2} Style}, or \textit{Architectural} style, is dated to around 80-20 BC. Paintings of this style often incorporated columns, windows, doors, and ledges, in order to create elaborate illusions of an extended space; intending to trick the viewers into thinking they are looking at an open, almost three-dimensional scene, instead of a wall.
    \item \textbf{\nth{3} Style}, or \textit{Ornamental} style, is dated to around 20 BC-50 AD. Simpler than the \nth{2} style, paintings of this style used broad, monochromatic planes of colors, while incorporating fantastic and stylized columns with fine details, including relatively simple scenes and creatures. 
    \item \textbf{\nth{4} Style}, or \textit{Intricate} style, is dated to around 40/50-79 AD. This style is often described as a combination of the previous three styles, incorporating elements from all of them.
\end{itemize}
Samples of frescos of each style are exemplified in Fig \ref{fig::styles}.

\subsection{POMPAAF Dataset Description}
\label{sec:datasets}

Not only do ancient Roman frescos provide us with a rare peek into the civilization and culture of roughly 2,000 years ago, but their four distinct artistic styles make an interesting test case for the art style classification task. Although preserved in relatively good shape, wall paintings in Pompeii still sustained damage and color fading, making the distinction between different art styles far from simple for the non-expert eye, even when attempted on full intact frescos. Moreover, the fact that the fourth style is a combination of visual features from all other three styles makes its identification, particularly in small fragments, significantly more challenging.

For that reason, we propose POMPAAF, a first of its kind dataset of style-annotated fragments of ancient roman wall paintings. The dataset is based on 51 images of \nth{1} style frescos, 103 images of \nth{2} style frescos, 78 images of \nth{3} style frescos, and 79 images of \nth{4} style frescos. All raw images are lossless scans of walls in Pompeii, which were sampled from the Pompeii Archive published by Napolitano \etal ~\cite{napolitano2023a}. 
The critical aspect of these published archives are manual annotations of all visual items that constitute these frescos, but none of these were used for our style classification work except the raw images themselves. 
It should also be noted that as the images were directly photographed from walls in the Pompeii Archaeological Park, their quality varies, both in terms of brightness and in terms of physical condition, as in some cases noticeable chunks of walls are missing.

\begin{figure}
    \centering
    \begin{tabular}{cccc}
        \includegraphics[width=0.22\columnwidth]{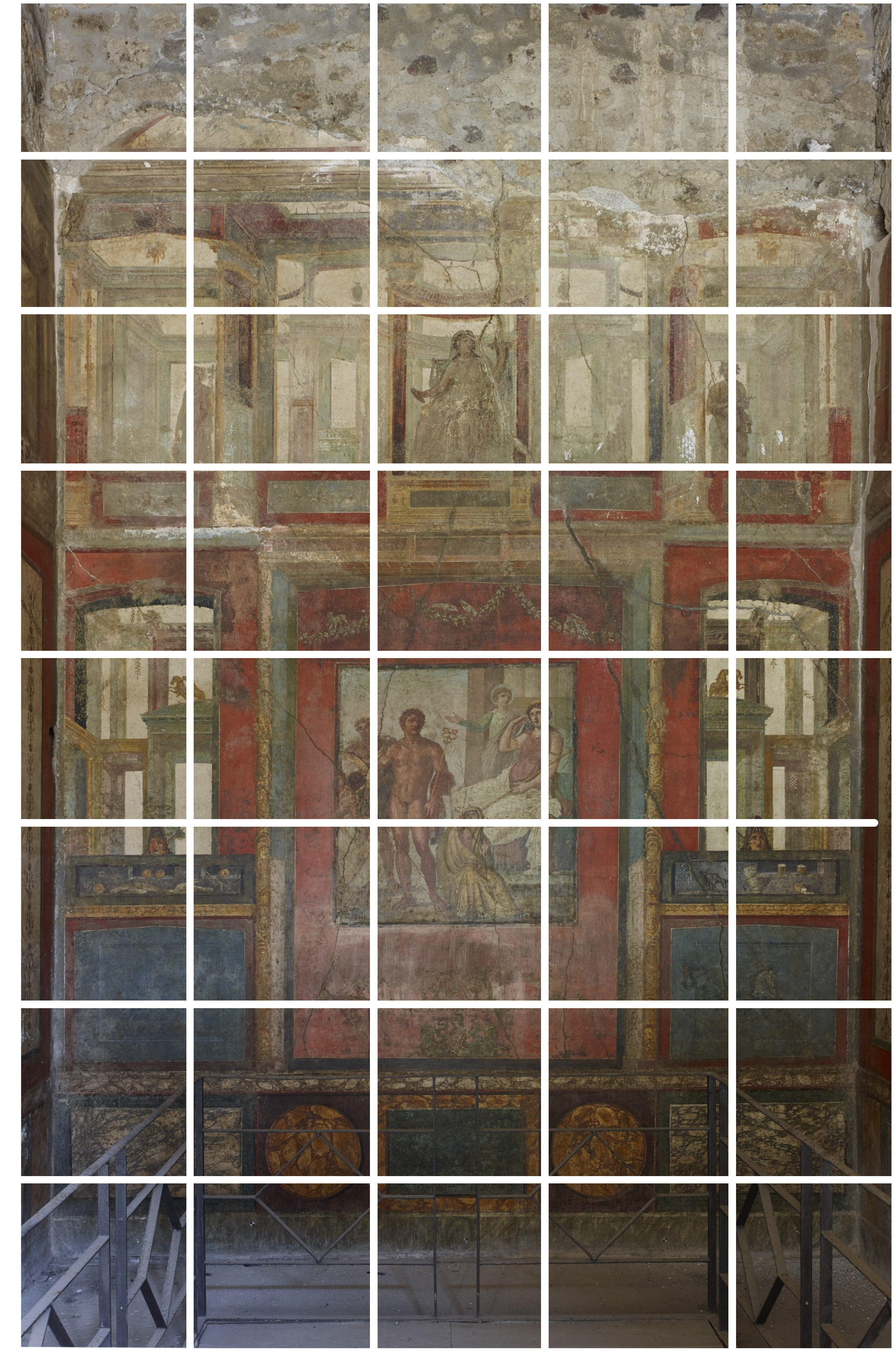}
        &
        \includegraphics[width=0.22\columnwidth]{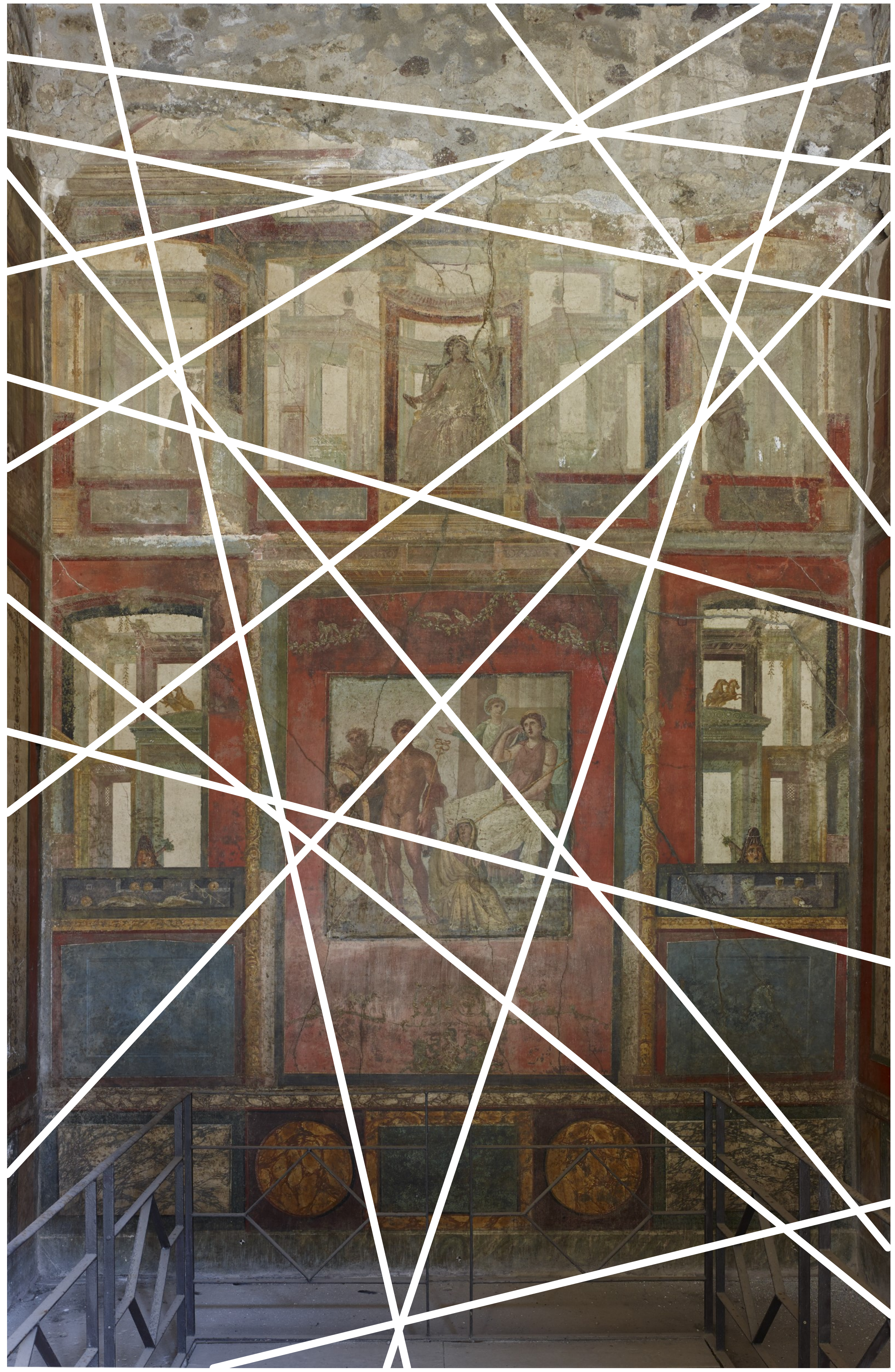}
        &
        \includegraphics[width=0.22\columnwidth]{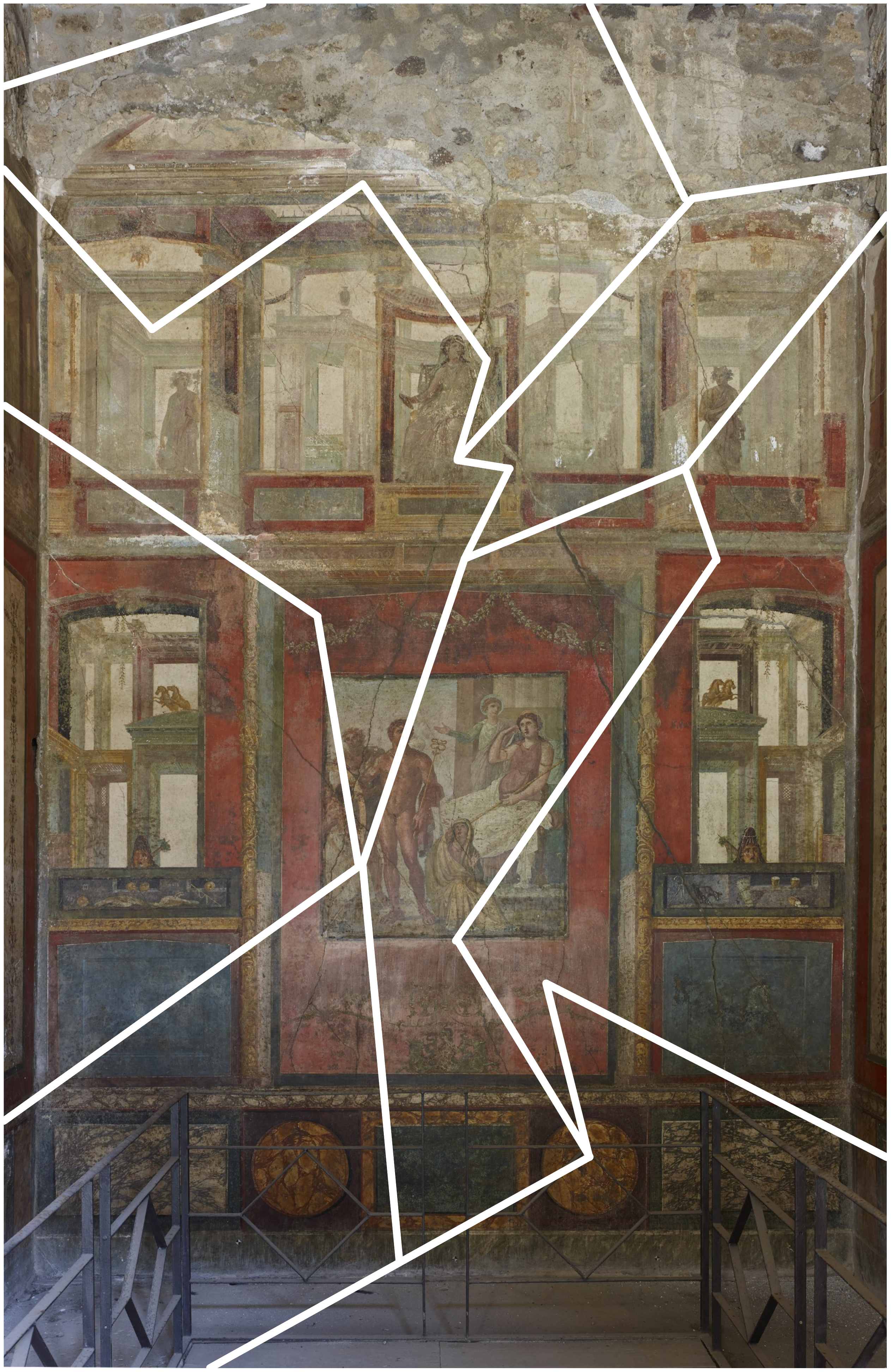}
        &
        \includegraphics[width=0.22\columnwidth]{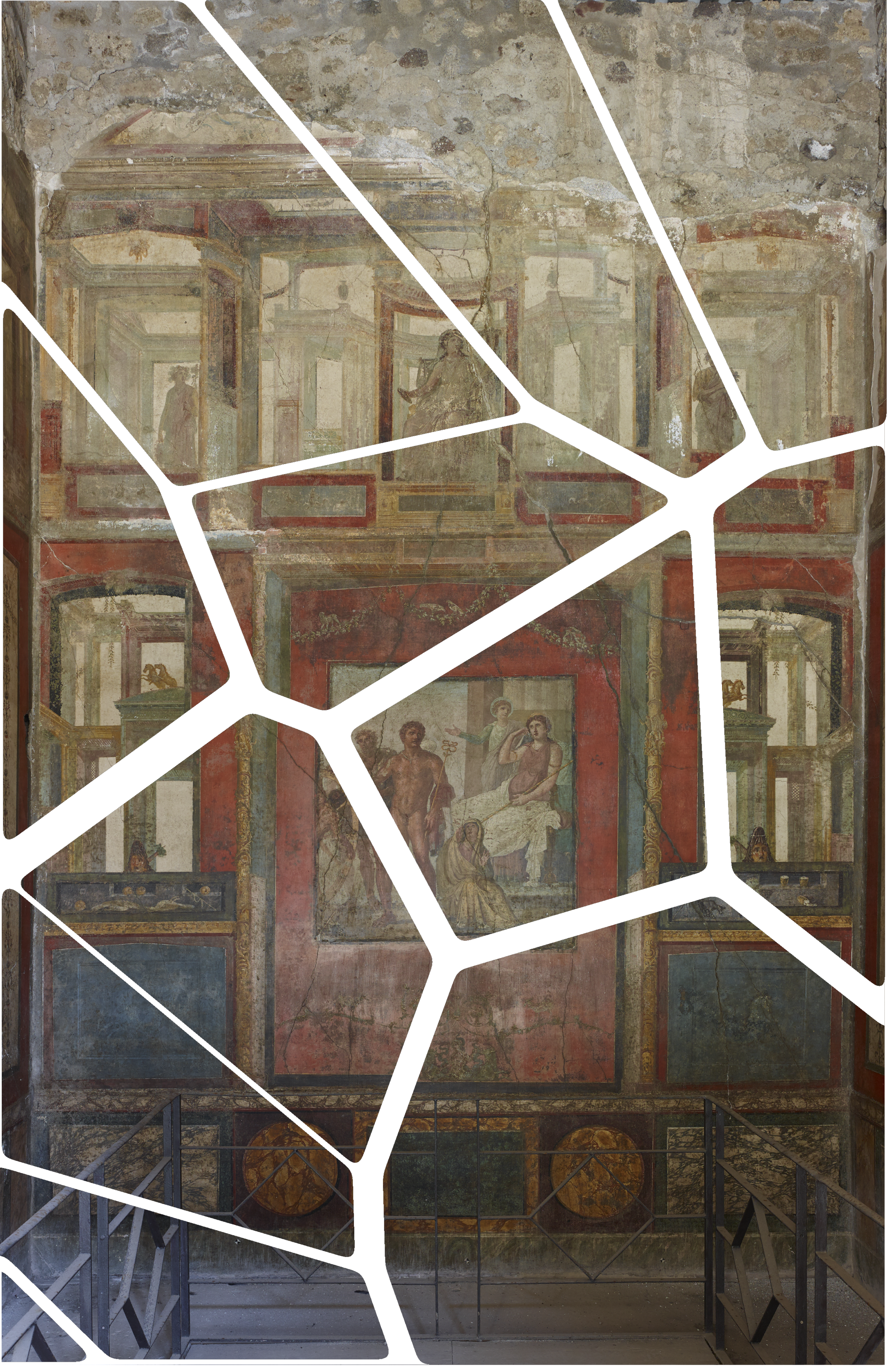}
        \\
        (a) & (b) & (c) & (d)
        \end{tabular}
    \caption{
           Examples of the four fragmentation methods used to create POMPAAF. The methods are demonstrated on a fresco of the \nth{4} Pompeian style, originally located in \textit{House of the Vettii}, Pompeii. (a) Strictly square fragments. (b) Crossing cuts polygonal fragments. (c) Non-convex partition fragments. (d) Eroded Voronoi partition frgments.}
  \label{fig::fragmentation_methods}
\end{figure}

To create POMPAAF, we artificially broke the fresco images into varying numbers of fragments, using four distinct methods, each generating a different geometrical pattern for the fragments, as described next and illustrated in fig \ref{fig::fragmentation_methods}. 

\begin{itemize}
    \item \textbf{Strictly Square Fragments.} For each fresco, we generated a different number of square fragments, all sharing the same dimensions and sizes, by dividing the fresco image, after slightly cropping it, into a pre-determined amount of square areas. All fragments were then rotated by a random degree. Each image was fragmented into 12, 40, 84, and 160 pieces. Of course, fragments of this type require the full image to be rectangular, as indeed we first cropped it from the original fresco image. This dataset is meant to be compatible with the vast square jigsaw puzzle literature~\cite{pomeranz2011fully,Paikin2015SolvingMS,andalo2016psqp,Son2019SolvingSJ}.
    
    \item \textbf{Crossing Cuts Polygonal Fragments.} Applying the method proposed by Harel et al.\cite{harel2024}, inspired by the ``Lazy Caterer'' sequence, we sampled a random convex polygon within each fresco image, then applied a varying number of \textit{crossing cuts} on it, breaking it into smaller convex polygons. Due to the stochastic nature of this method, it is impossible to directly control the number of fragments obtained for each fresco, thus we instead controlled the number of cuts; using 5, 10, 15, or 20 cuts, which generated approximated averages of 12, 42, 88, and 151 fragments, respectively.   
    
    \item \textbf{Non-convex Partition Fragments.} To divide an image into $N$ non-convex polygonal fragments, we start with a random Delaunay triangulation of $R>N$ points on the image plane, yielding $N'$ simplices. Then, we iteratively merge neighboring pieces until the desired number of fragments is achieved. Smaller fragments are prioritized for being merged first, to discourage high variance between fragments in the outcome puzzle. When two fragments are merged, their joint contour can become non-convex, thus providing this process its name. Each image was fragmented into 12, 40, 84, and 160 pieces.
    
    \item \textbf{Eroded Voronoi Partition Fragments.} For this dataset we sampled $N$ points on the image plane and calculated their Voronoi cells using a nearest-neighbor approach. Then, we simulated erosion by retracting a random number of pixels (up to 30) from the cells' boundaries, followed by a smoothing operation on the fragments' contours. This process was designed to yield natural-looking eroded fragments, similar to those found in the DAFNE dataset \cite{dondi2020dafne}. Each image was fragmented into 12, 40, 84, and 160 pieces. 
\end{itemize}

\section{Style Identification and Classification}
\label{sec:method}

\begin{figure}
    \centering
    \includegraphics[width=0.9\textwidth]{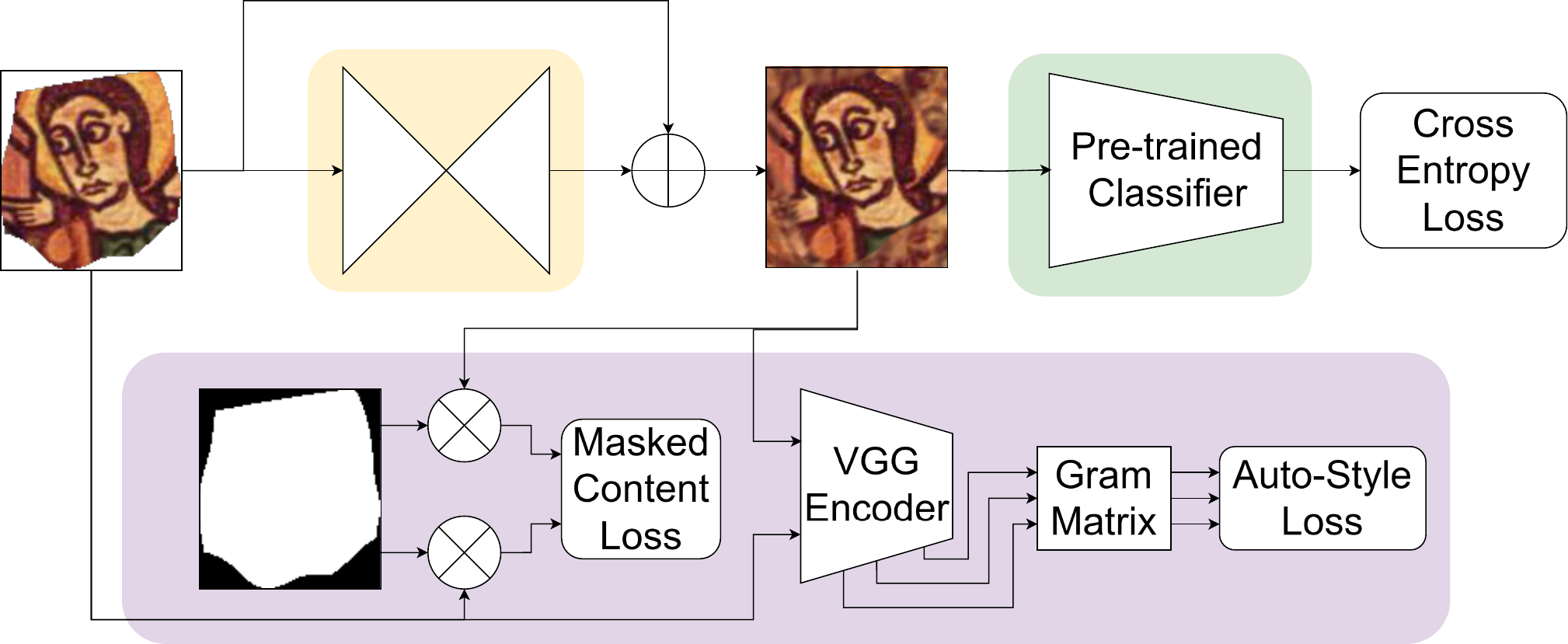}
    \caption{Diagram of the proposed model. \textbf{Yellow (top-left):} Style extrapolation module. \textbf{Green (top-right):} Pre-trained image classifier. \textbf{Purple (bottom):} Loss computation for the style extrapolator. The depicted fragment is from the CLEOPATRA \cite{cascone2023classification} training set.}
    \label{fig:model_diag}
\end{figure}

Here we describe our generic deep-learning framework for style classification, designed to be robust in the face of image fragmentation. Our architecture, illustrated in \cref{fig:model_diag}, comprises two main tasks:
\begin{itemize}
    \item \textbf{Style Extrapolation.} This preprocessing module combines several loss functions over a modified convolutional autoencoder to overcome the problems that emerge when handling \textit{fragments} of images (as further explained in \cref{sec:extrapol_module}). This unique component and loss functions constitute the primary innovation of our approach, deviating from traditional style recognition methods.
    \item \textbf{Style Classification.} The main classifier utilizes transfer learning with top-scoring image classification architectures for accurate and efficient style recognition.
\end{itemize}
The whole system is optimized through a two-step process -- first, the style extrapolator is trained using its hand-crafted losses (see \cref{sec:extrapol_module}). Then it is frozen and used for transforming the classification module's inputs during training and inference.

In the style transfer community, the \textit{style} of a picture is considered independent of its \textit{content} (see \cref{sec:related}). However, in archaeological research, ``artistic style'' often involves content-based features such as certain characters or objects, eminent from \cref{sec:data}. To overcome this conflict when learning to recognize archaeological style, our network is designed to capture both types of features---the style-transfer-based extrapolator expands the pure style of the image while preserving its content; the object-detection-trained classification module can then recognize both kinds of elements.

\subsection{Style Extrapolation Module}
\label{sec:extrapol_module}

When considering images of fresco fragments, the geometry of the piece's contour plays a non-negligible role in the pictorial content of the full image, as all pixels outside of the contour represent the background (typically by being assigned some homogeneous color). Indeed, this sort of representation becomes one of the main challenges for accurate classification when handling partial and irregularly shaped input. To reduce this effect, some pre-processing must be done. One approach is to ignore or discard the non-pictorial pixels, which could impair the spatial arrangement of the classifier's input. Alternatively, one could try to extrapolate the piece to fill the whole image domain, but this approach might introduce ``out-of-style'' elements to the image. As an intermediate approach, we designed a differentiable fragment pre-processing module, aimed at ``diffusing'' the fragments's stylistic features into the whole image domain.

This style extrapolation module exhibits a modified autoencoder architecture, where the encoder part utilizes a pre-trained ResNet18 \cite{he2016deep}, and the decoder extracts the compressed representation back to the original image dimensions using several convolutional layers. We introduce a residual connection, aimed to simplify the preservation of critical input features by summing the network's input and output. The component's end goal is to generate a pictorially full and stylistically identical version of its input, and it is optimized by a specially curated loss function, inspired by neural style transfer. 

Building on top of traditional style loss, presented in \cref{eq:style_loss_orig}, we define the auto-style loss:
\begin{equation}
    \mathcal{L}_\mathrm{AS}=\sum_\ell \frac{w_\ell}{4m_\ell^2n_\ell^2}\Vert\mathrm{Gram}(I_\ell)-\mathrm{Gram}(E(I)_\ell)\Vert_F^2
    \label{eq:auto_style_loss}
\end{equation}
where $E(I)$ is the extrapolator's output for $I$, which is identical in dimensions to its input. Note that to calculate this loss we pass $I$ and $E(I)$ through an external pre-trained VGG \cite{simonyan2014very}. As described in \cref{sec:related}, the Gram matrix computes the correlation between different components of a feature map, while disregarding spatial positions. Because stylistic features are represented both in higher and lower levels of abstraction, we obtain an overall stylistic similarity measure by comparing the Gram matrices from various depths of the VGG, as developed by Gatys \etal \cite{gatys2016image}. In Principle, this loss function guides the style extrapolation module to preserve the style of its input as much as possible. 

Additionally, we present a masked version of the content loss from \cref{eq:content_loss_orig}
\begin{equation}
    \mathcal{L}_\mathrm{MC} = \frac{1}{2}\Vert M\odot(I-E(I))\Vert_F^2
\end{equation}
where $\odot$ denotes element-wise multiplication. The mask $M$ should assign values that are closer to 1 for fragment pixels and approaching 0 values for background pixels. A natural choice would be to use the image's alpha channel.
Finally, combining both losses, we obtain our style extrapolation loss:
\begin{equation}
    \mathcal{L}_\mathrm{SE}=\lambda\mathcal{L}_\mathrm{AS}+\mu\mathcal{L}_\mathrm{MC} \;\;.
\end{equation}

\subsection{Style Classification Module}
\label{sec:class_module}

Artistic style is defined by both low-level features, such as color and composition, as well as high-level features, such as techniques and recurring themes. This complexity makes the task particularly suitable for CNNs, which excel at automatically extracting and learning features at multiple levels of abstraction. 
While other choices can be made, here we employed an EfficientNet \cite{tan2019efficientnet} model as our classifier backbone, pre-trained on ImageNet \cite{deng2009imagenet} classification. This choice is motivated in part by EfficientNet's ability to balance model size and accuracy, ensuring efficient feature extraction and robust performance.

It is commonly agreed upon that the low-level features of the image are extracted at the more primary convolutional layers, and they become more high-level and semantic at the deeper layers. Because our backbone was pre-trained on categorizing the \textit{content} of its input image rather than its \textit{style}, its high-level features might not be ideally suited for our needs. Nonetheless, the low-level features are presumably common to both tasks, as they are argued to suit most (and possibly all) types of natural images. Hence, we use only the primary layers of our backbones as-is (frozen) and fine-tune the other layers by proper training using standard categorical cross-entropy loss:
\begin{equation}
    \mathcal{L}_\mathrm{CE} = -\sum_{k=1}^K y_k\log(p_k)\;,
\end{equation}
where $y_k$ indicates if the input is of style $k$ (within $K$ styles) and $p_k$ is the network's assigned probability for the input to be an instance of that style.

\subsection{Baselines}

To demonstrate the utility of the proposed approach, we of course compared it to the prior art, but also against two additional baselines, whose utility is twofold. First, because to our best knowledge no code is readily available for comparing the previous SOTA by Cascone \etal~\cite{cascone2023classification} on novel datasets (like the ones in POMPAAF). And second, because it was likely that even naive deep learning methods could match that traditional SOTA. The two baselines we  suggest are thus:
\begin{itemize}
    \item \textbf{CNN}: We designed a simple CNN, composed of two convolutional blocks with ReLU activation and max-pooling. The output of the second block is passed to a linear layer for multi-class prediction. This baseline was tested to demonstrate the potential of a simplistic yet lightweight model; and to serve as a point of reference.
    \item \textbf{Transfer Learning (TL)}: In order to show the necessity of the style extrapolation module, we examined the performance of a stand-alone pre-trained EfficientNet \cite{tan2019efficientnet} classifiers. 
\end{itemize}

\section{Results}
\label{sec:results}
Here we present the test scores of various models on the existing CLEOPATRA dataset \cite{cascone2023classification} and the proposed POMPAAF datasets. Furthermore, we use the multiple fragmentation types from the POMPAAF dataset to test the effect of the fragment's geometry on all architectures. Moreover, the incremental growth in the number of pieces over the different subsets of POMPAAF allows for a closer look at a possible trade-off between data quality and quantity, further elaborated in \cref{sec:POMPAAF_res}.

\subsection{Evaluation Metrics}
We evaluate the aforementioned models using standard classification metrics:
\begin{equation}
\label{eq:accuracy}
    \mathrm{accuracy}(s_k)=\frac{TP+TN}{TP+TN+FP+FN}
\end{equation}
\begin{equation}
\label{eq:precision}
    \mathrm{precision}(s_k)=\frac{TP}{TP+FP}
\end{equation}
\begin{equation}
\label{eq:recall}
    \mathrm{recall}(s_k)=\frac{TP}{TP+FN}
\end{equation}
\begin{equation}
\label{eq:f1}
    \mathrm{F1}=2\frac{\mathrm{precision}\cdot\mathrm{recall}}{\mathrm{precision}+\mathrm{recall}}
\end{equation}
where $s_k$ denotes a specific style, and $TP$ (true positive) for example counts the number of fragments that were correctly classified as ``style $k$''. In principle, \textit{accuracy} computes the number of correct predictions out of the total; \textit{precision} measures the ratio of samples classified as ``style $k$'' that indeed belong to style $k$; \textit{recall} calculates the fraction of actual style $k$ instances that were correctly identified; and lastly, the \textit{F1 score} is the harmonic mean of precision and recall. We used macro-averaging to generalize the metrics in \cref{eq:precision,eq:recall} for multi-style recognition.

\subsection{CLEOPATRA Challenge}
\label{sec:cleo_res}
First, we demonstrate our results on the CLEOPATRA dataset \cite{cascone2023classification}, where the SOTA was based on a random forest model \cite{breiman2001random}.

\begin{table}
\centering
\caption{Results on the CLEOPATRA Test-Set (80 pieces) \cite{cascone2023classification}}
\begin{tabular}{lS[table-format=1.3]ccc}
\toprule
\multirow{2}{*}{\textbf{Metric}} & \textbf{Cascone} & \textbf{CNN} & \textbf{TL} & \textbf{Proposed} \\
 & \textbf{~\etal\cite{cascone2023classification}} & \textbf{~baseline~} & \textbf{~baseline~} & \textbf{~method} \\
\midrule
Accuracy $\uparrow$      & 0.28             & 0.265                 & 0.406                & \textbf{0.475}           \\
Precision $\uparrow$     & 0.29             & 0.277                 & 0.411                & \textbf{0.459}           \\
Recall $\uparrow$        & 0.253*           & 0.264                 & 0.406                & \textbf{0.473}           \\
F1 $\uparrow$            & 0.27             & 0.270                 & 0.408                & \textbf{0.466}           \\
\bottomrule
\end{tabular}\\
\footnotesize{*Not provided in Cascone \etal~\cite{cascone2023classification}, but calculated from their precision and F1 scores}
\label{tab:results_cleopatra}
\end{table}

As evident in \cref{tab:results_cleopatra}, the computational expressiveness of modern architectures outperforms the traditional models~\cite{cascone2023classification} as well as the CNN baseline. More importantly, the non-negligible gap between our proposed model and the transfer-learned one testifies to the benefits of including the style extrapolation module.

\subsection{POMPAAF Dataset}
\label{sec:POMPAAF_res}

Although the CLEOPATRA dataset \cite{cascone2023classification} is extensive and diverse, we opted to test our method on a more operative scenario, where style classification is required on fragments that are found in the same site but can be from different styles if they belonged to different frescos (painted on different walls). In practice, when frescos from different walls do mix up, their styles can be very close, both semantically and pictorially. This phenomenon is present in the Pompeian archive dataset, where styles 1-4 are similar in many visual features, but still recognized as different (cf. Sec.~\ref{sec:four_styles}).

The results of our proposed model against other baselines on all subsets of the POMPAFF dataset are found in \cref{tab:res_square,tab:res_cc,tab:res_ncp,tab:res_vor}. Evidently, our architecture outperforms all other baselines by a far margin, serving as a new point of reference for classifying the artistic styles of cultural heritage pictorial fragments. Also clear is the better overall performance compared to \cref{tab:results_cleopatra}. Indeed, even though the similarity between the four Pompeian styles is significant, the relatively small variance between instances of the same style as well as the smaller number of classes, help drive general performance up.

When considering the number of pieces into which the original image was divided, an interesting trade-off arises. With a larger number of fragments, each piece contains less pictorial information. However, more pieces also provide more training examples. One could thus argue that with smaller but more numerous fragments, the total amount of information fed to the network remains stable, while its quality deteriorates, and thus performance is likely to decline. To investigate this, we conducted several experiments, incrementally increasing the number of pieces each time. Surprisingly, as evident from \cref{tab:res_square,tab:res_cc,tab:res_ncp,tab:res_vor}, an opposite pattern emerges -- more training examples of smaller fragments in fact \textit{improves} accuracy across different models. This phenomenon likely results from the observation that most architectures rely on a fixed input size with a predefined resolution. For example, our pre-trained EfficientNet \cite{tan2019efficientnet} expects images of $224\times 224$ pixels. Hence, larger pieces must be downscaled more aggressively, thus losing some visual features. In other words, smaller pieces, although covering less area from the original picture, actually preserve more subtle information overall, leading to better scoring models.

Another interesting experiment is testing the effect of the fragmentation method (\ie fragments' shape) on the model's accuracy. Comparing the results of the same models on different fragmentation techniques, some phenomena become apparent:
\begin{itemize}

    \item All tested models perform best over the strictly square fragments. This result is not surprising because the square shape matches the typical input of convolutional networks, and no background regions of arbitrary shape are created to confound the classification process.
    
    \item Compared to the other fragmentation methods, models training on the crossing-cuts dataset tend to suffer from relatively low accuracy. A likely reason for that phenomenon is the high variance in piece size for crossing-cuts fragmentation, as revealed by statistical analysis of the different fragment area distributions. Specifically, square pieces are all identical in size and shape, while the non-convex partition algorithm prioritizes the merging of smaller fragments first, leading to more similar piece sizes. Models training to classify crossing-cuts fragments, however, need to handle both scarce details in minuscule fragments next to large-scale content in extensive regions of the original image, a significantly more challenging task. This hypothesis was put to the test by discarding the prioritization mechanism of the non-convex partition algorithm (cf.~\cref{sec:datasets}) and re-running all models on the new data. When the pieces are merged randomly with no consideration of size, the result is a larger deviation in fragment sizes, similar to the corssing-cuts puzzles. Under these conditions we observed a consistent decrease in accuracy of about 8\% by all models, confirming the hypothesis.
    
\end{itemize}

\begin{table}
\centering
\caption{Results on the \textbf{Strictly Square} test set}
\begin{tabular}{clccc}
\toprule
\textbf{~~\#Pieces~~}    & \textbf{Metric}   & \textbf{~CNN~} & \textbf{~~~TL~~~} & \textbf{Proposed} \\
\midrule
\multirow{4}{*}{12}  & Accuracy  & 0.548                   & 0.627                                 & \textbf{0.846}                \\
                     & Precision & 0.559                   & 0.624                                 & \textbf{0.840}                 \\
                     & Recall    & 0.542                   & 0.636                                 & \textbf{0.863}                \\
                     & F1        & 0.550                   & 0.630                                 & \textbf{0.851}                \\
\midrule
\multirow{4}{*}{40}  & Accuracy  & 0.574                   & 0.638                                 & \textbf{0.910}                 \\
                     & Precision & 0.579                   & 0.635                                 & \textbf{0.908}                \\
                     & Recall    & 0.569                   & 0.649                                 & \textbf{0.917}                \\
                     & F1        & 0.574                   & 0.642                                 & \textbf{0.913}                \\
\midrule
\multirow{4}{*}{84}  & Accuracy  & 0.614                   & 0.627                                 & \textbf{0.949}                \\
                     & Precision & 0.614                   & 0.622                                 & \textbf{0.946}                \\
                     & Recall    & 0.611                   & 0.640                                 & \textbf{0.955}                \\
                     & F1        & 0.612                   & 0.631                                 & \textbf{0.950}                \\
\midrule
\multirow{4}{*}{160} & Accuracy  & 0.640                   & 0.634                                 & \textbf{0.967}                \\
                     & Precision & 0.638                   & 0.634                                 & \textbf{0.969}                \\
                     & Recall    & 0.638                   & 0.641                                 & \textbf{0.971}                \\
                     & F1        & 0.638                   & 0.637                                 & \textbf{0.970}                \\
\bottomrule
\end{tabular}
\label{tab:res_square}
\end{table}

\begin{table}
\centering
\caption{Results on the \textbf{Crossing-Cuts} test set}
\begin{tabular}{clccc}
\toprule
\textbf{~~\#Cuts~~}      & \textbf{Metric}   & \textbf{~CNN~} & \textbf{~~~TL~~~} & \textbf{Proposed} \\
\midrule
\multirow{4}{*}{5}  & Accuracy  & 0.417                    & 0.556                                  & \textbf{0.720}              \\
                    & Precision & 0.408                    & 0.557                                  & \textbf{0.717}              \\
                    & Recall    & 0.393                    & 0.556                                  & \textbf{0.727}              \\
                    & F1        & 0.400                    & 0.556                                  & \textbf{0.722}              \\
\midrule
\multirow{4}{*}{10} & Accuracy  & 0.508                    & 0.531                                  & \textbf{0.783}              \\
                    & Precision & 0.500                    & 0.531                                  & \textbf{0.783}              \\
                    & Recall    & 0.475                    & 0.529                                  & \textbf{0.786}              \\
                    & F1        & 0.487                    & 0.530                                  & \textbf{0.784}              \\
\midrule
\multirow{4}{*}{15} & Accuracy  & 0.518                    & 0.538                                  & \textbf{0.807}              \\
                    & Precision & 0.509                    & 0.536                                  & \textbf{0.804}              \\
                    & Recall    & 0.505                    & 0.540                                  & \textbf{0.820}              \\
                    & F1        & 0.507                    & 0.538                                  & \textbf{0.812}              \\
\midrule
\multirow{4}{*}{20} & Accuracy  & 0.535                    & 0.526                                  & \textbf{0.821}              \\
                    & Precision & 0.530                    & 0.525                                  & \textbf{0.822}              \\
                    & Recall    & 0.523                    & 0.524                                  & \textbf{0.826}              \\
                    & F1        & 0.526                    & 0.524                                  & \textbf{0.824}              \\  
\bottomrule
\end{tabular}
\label{tab:res_cc}
\end{table}

\begin{table}
\centering
\caption{Results on the \textbf{Non-convex Partition} test set}
\begin{tabular}{clccc}
\toprule
\textbf{~~\#Pieces~~}    & \textbf{Metric}   & \textbf{~CNN~} & \textbf{~~~TL~~~} & \textbf{Proposed} \\
\midrule
\multirow{4}{*}{12}  & Accuracy  & 0.459                   & 0.597                                 & \textbf{0.867}                \\
                     & Precision & 0.45                    & 0.595                                 & \textbf{0.862}                \\
                     & Recall    & 0.445                   & 0.606                                 & \textbf{0.872}                \\
                     & F1        & 0.447                   & 0.600                                 & \textbf{0.867}                \\
\midrule
\multirow{4}{*}{40}  & Accuracy  & 0.541                   & 0.601                                 & \textbf{0.904}                \\
                     & Precision & 0.531                   & 0.604                                 & \textbf{0.907}                \\
                     & Recall    & 0.538                   & 0.596                                 & \textbf{0.904}                \\
                     & F1        & 0.534                   & 0.600                                 & \textbf{0.905}                \\
\midrule
\multirow{4}{*}{84}  & Accuracy  & 0.564                   & 0.578                                 & \textbf{0.925}                \\
                     & Precision & 0.565                   & 0.578                                 & \textbf{0.923}                \\
                     & Recall    & 0.55                    & 0.584                                 & \textbf{0.93}                 \\
                     & F1        & 0.557                   & 0.581                                 & \textbf{0.926}                \\
\midrule
\multirow{4}{*}{160} & Accuracy  & 0.602                   & 0.583                                 & \textbf{0.945}                \\
                     & Precision & 0.61                    & 0.582                                 & \textbf{0.945}                \\
                     & Recall    & 0.612                   & 0.586                                 & \textbf{0.948}                \\
                     & F1        & 0.611                   & 0.584                                 & \textbf{0.946}                \\
\bottomrule
\end{tabular}
\label{tab:res_ncp}
\end{table}

\begin{table}
\centering
\caption{Results on the \textbf{Eroded Voronoi Partition} test set}
\begin{tabular}{clccc}
\toprule
\textbf{~~\#Pieces~~}    & \textbf{Metric}   & \textbf{~CNN~} & \textbf{~~~TL~~~} & \textbf{Proposed} \\
\midrule
\multirow{4}{*}{12}  & Accuracy  & 0.527                   & 0.581                                 & \textbf{0.846}                \\
                     & Precision & 0.528                   & 0.587                                 & \textbf{0.843}                \\
                     & Recall    & 0.517                   & 0.590                                 & \textbf{0.852}                \\
                     & F1        & 0.522                   & 0.588                                 & \textbf{0.847}                \\
\midrule
\multirow{4}{*}{40}  & Accuracy  & 0.555                   & 0.592                                 & \textbf{0.874}                \\
                     & Precision & 0.554                   & 0.596                                 & \textbf{0.876}                \\
                     & Recall    & 0.551                   & 0.591                                 & \textbf{0.881}                \\
                     & F1        & 0.552                   & 0.593                                 & \textbf{0.878}                \\
\midrule
\multirow{4}{*}{84}  & Accuracy  & 0.565                   & 0.579                                 & \textbf{0.873}                \\
                     & Precision & 0.558                   & 0.586                                 & \textbf{0.876}                \\
                     & Recall    & 0.567                   & 0.574                                 & \textbf{0.873}                \\
                     & F1        & 0.562                   & 0.579                                 & \textbf{0.874}                \\
\midrule
\multirow{4}{*}{160} & Accuracy  & 0.551                   & 0.567                                 & \textbf{0.876}                \\
                     & Precision & 0.561                   & 0.568                                 & \textbf{0.888}                \\
                     & Recall    & 0.547                   & 0.562                                 & \textbf{0.871}                \\
                     & F1        & 0.553                   & 0.564                                 & \textbf{0.879}                \\
\bottomrule
\end{tabular}
\label{tab:res_vor}
\end{table}

\section{Conclusion}
\label{sec:conclusion}

We presented a novel architecture for addressing style classification of pictorial cultural heritage fragments. Our approach employed modern deep learning architectures, optimized with loss functions that preserve stylistic elements while overcoming missing parts in the fragment's image. This combined method yields SOTA results on both the CLEOPARTA \cite{cascone2023classification} and the POMPAAF datasets. As image classification accuracy continues to improve, we anticipate future studies on fragment style recognition will surpass this new baseline by utilizing the growing power of contemporary architectures and domain-specific knowledge. In particular, with recent advancements in attention-based models for image classification \cite{dosovitskiy2020image,liu2021swin} redefining the field, their performance on our task would be worth exploring. Furthermore, with the increasing popularity of diffusion models for image generation and inpainting \cite{lugmayr2022repaint}, combined with the diverse nature of style losses \cite{li2017demystifying}, future style extrapolation modules could take many forms and advance even more.
\newpage

\begin{credits}
\subsubsection{\ackname} 
This work has been funded in part by the European Union’s Horizon 2020 research and innovation programme under grant agreement No 964854 (the RePAIR project). We also thank the Helmsley Charitable Trust through the ABC Robotics Initiative and the Frankel Fund of the Computer Science Department at Ben-Gurion University for their generous support.

\subsubsection{\discintname} The authors have no competing interests to declare that are relevant to the content of this article. 
\end{credits}

\bibliographystyle{splncs04}
\bibliography{main}

\end{document}